\documentclass{article}
\usepackage{spconf,amsmath,graphicx}
\usepackage{tikz}
\usepackage{pgfplots}
\usepackage{caption}
\usepackage{multirow}
\usepackage{url}
\title{Brain Age Estimation Using LSTM on Children's Brain MRI}
 \name{\begin{tabular}{c}Sheng He$^{\star}$  \qquad  Randy L. Gollub$^\dagger$  \qquad  Shawn N. Murphy$^\dagger$ \qquad Juan David Perez$^\star$  \qquad Sanjay Prabhu$^\star$ \qquad \\  Rudolph Pienaar$^\star$ \qquad Richard L. Robertson$^\star$ \qquad P. Ellen Grant$^\star$ \qquad Yangming Ou$^\star$ \end{tabular}}

 \address{$^{\star}$ Boston Children's Hospital, Harvard Medical School, Boston , USA\\
     $^{\dagger}$ Massachusetts  General  Hospital, Harvard Medical School, Boston, USA}
\begin{document}
%\ninept
%
\maketitle
\begin{abstract}
Brain age prediction based on children's brain MRI is an important biomarker for brain health and brain development analysis.
In this paper, we consider the 3D brain MRI volume as a sequence of 2D images and propose a new framework using the recurrent neural network for brain age estimation.
The proposed method is named as 2D-ResNet18+Long short-term memory (LSTM), which consists of four parts:
2D ResNet18 for feature extraction on 2D images, a pooling layer for feature reduction over the sequences, an LSTM layer, and a final regression layer.
We apply the proposed method on a public multisite NIH-PD dataset and evaluate generalization on a second multisite dataset,
which shows that the proposed 2D-ResNet18+LSTM method provides better results than traditional 3D based neural network for brain age estimation.

\end{abstract}
\begin{keywords}
MRI, Age Prediction, ResNet, LSTM
\end{keywords}
\section{Introduction}
\label{sec:intro}

Predicting brain age from brain MRI is becoming an important biomarker for brain health and brain development analysis~\cite{cole2017brain,cole2017predicting}.
The difference between the predicted and chronological age can be used to predict neurocognitive disorders~\cite{aycheh2018biological} or brain anomaly caused by disease~\cite{cole2017predicting}.

Inspired by deep learning, 3D-Convolutional Neural Network (3D-CNN) has been used to predict brain ages from a 3D brain MRI~\cite{ueda2019age}. However, 3D-CNN requires extensive computations and more memory than 2D-CNN and it is hard to parallelize computations.
%and training a 3D-CNN with a small dataset is prone to over-fitting. 
A 2017 study used 3D-CNN on MRI to predict brain ages of N=2001 subjects aged from 18-90 years and reported an average prediction error of 4.16 years~\cite{cole2017predicting}. We are particularly interested in 0-20 years especially 0-6 years of age, where data is hard to get, the sample size is smaller, the brain is rapidly developing, and the inter-subject variability is bigger.
Because of all these, 3D-CNN faces an additional risk for over-fitting.

To solve these problems, we propose a novel framework which considers the 3D brain MRI volume as a sequence of 2D images and uses recurrent neural network (such as the typical Long Short-Term Memory or LSTM~\cite{hochreiter1997long}) to capture age information over the sequence of 2D slices. A similar approach (treating a 3D MRI as a series/sequence of 2D slices) was recently developed for voxel-wise abnormality detection~\cite{zhu2018exploiting}, but we face a different task -- patient-level prediction rather than voxel-wise classification, and hence a different formulation. In our framework, the feature representations of 2D slices are extracted by 2D-CNN which is easy to be parallelized and they are fed into LSTM to capture the global age information, which we assume is less likely to be trapped at local minima and less likely to over-fit in a small sample size in this age range.

\begin{figure}[!t]
	\centering
	\includegraphics[width=0.5\textwidth]{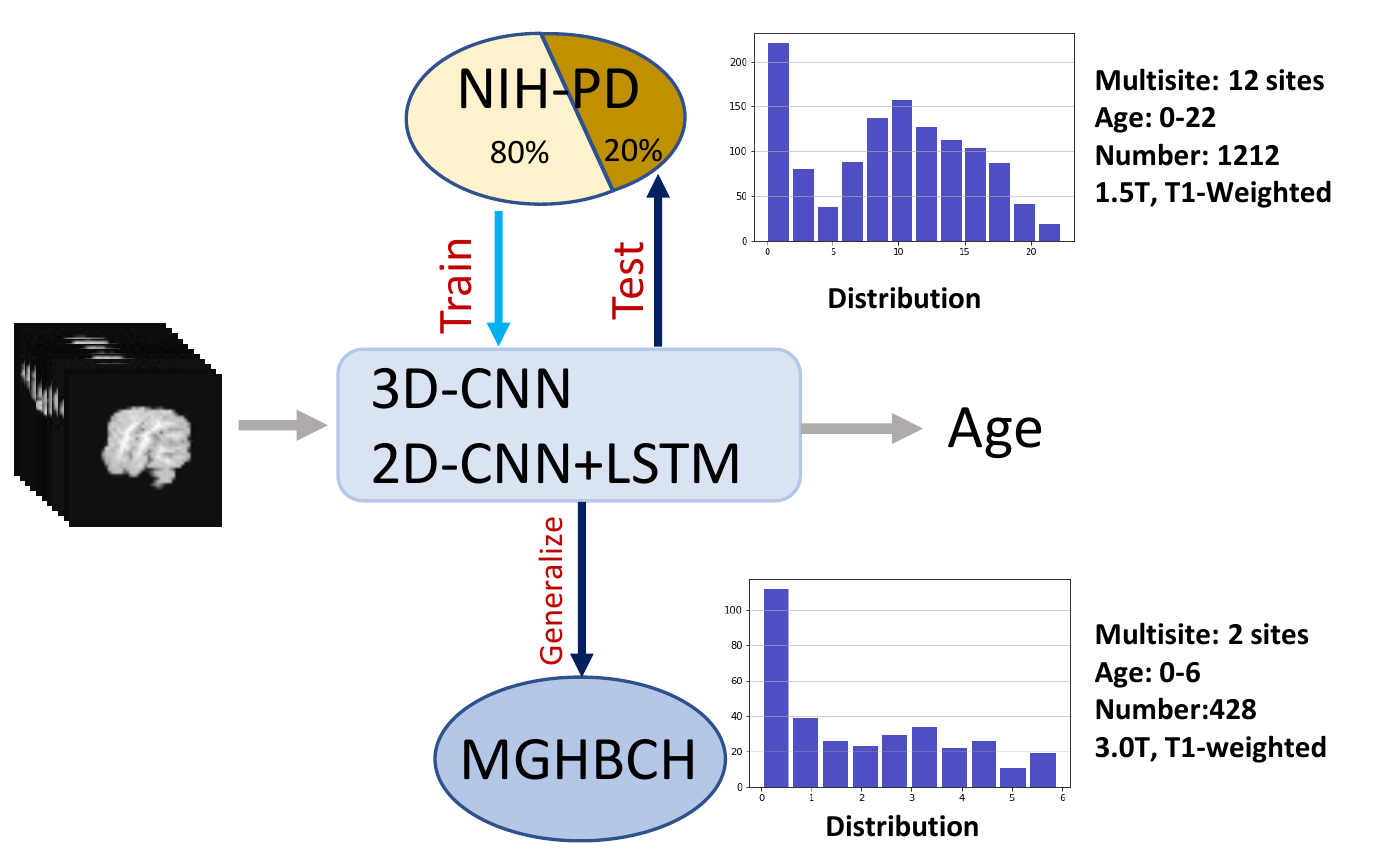}
	\caption{The proposed framework for age estimation.
	Using NIH-PD as the Discovery Cohort, we train two different models (the typical 3D-CNN and the proposed 2D-CNN+LSTM) on the training set (80\% of NIH-PD data) and test on the testing set (20\% of NIH-PD data). Then, using MGHBCH as the Replication Cohort, we quantify the generalization ability across datasets.}
	\label{fig:model}
\end{figure}

In most brain age estimation studies, all available data are split into training and testing sets. Some of the training and testing subjects share imaging sites/protocols. So, although the exact testing subjects are unseen during training, the imaging sites/protocols may have already been exposed to training. In contrast, we use two multisite datasets: the NIH-Pediatric Data (NIH-PD)~\cite{evans2006nih}~\footnote{\url{https://www.nitrc.org/forum/forum.php?forum\_id=1152}} (GE/Siemens 1.5T scanner, 12 sites, spin-echo T1-weighted MRI), and the MGHBCH dataset (Massachusetts General and Boston Children's Hospitals, Siemens 3.0T Trio scanners, T1-weighted MPRAGE sequence). The two datasets have very different imaging sites/protocols, and by training/testing on NIH-PD (the Discovery Cohort) and replicating on MGHBCH data (the Replication Cohort, Fig~\ref{fig:model}), we fully evaluate 3D-CNN versus our proposed algorithm for the generalization ability in truly "unseen" dataset --  not only testing subjects unseen but their imaging protocols/sites unseen.

%The framework is shown in Fig.~\ref{fig:model}. The NIH-PD dataset is split into training (80\%) and testing (20\%) subsets. The model is trained using the training set and tested on the testing set. The generalization ability is evaluated on the MGHBCH dataset. %Note that the age range (0-6) of the MGHBCH is a subset of the age range of the NIH-PD (0-20). % We use two models for age estimation: the proposed 2D-CNN+LSTM and the typical 3D-CNN for comparison.

%It contains T1-weighted MPRAGE (TR=~2.5s, TE=~5.34ms, TI=~1100ms, 1mm isotropic voxel size).

%thus the generalization ability which is the results on a dataset by using the model trained from another different dataset is unknown.

%We evaluate the proposed method on children's brain MRI from newborn to 20 years old on the NIH-Pediatric Data (NIH-PD)~\cite{evans2006nih}~\footnote{\url{https://www.nitrc.org/forum/forum.php?forum_id=1152}}, which is a multisite dataset and brain MRIs are collected from 12 different institutes. 

%For the generalization evaluation, we apply the trained model on the NIH-PD dataset on a new collection: the MGHBCH dataset which contains brain MRIs of 428 children's 0-6 years of age.
%The MGHBCH dataset is collected from Massachusetts General Hospital and Boston Children's Hospital, by a Siemens 3.0T Trio scanner. 
%It contains T1-weighted MPRAGE (TR=~2.5s, TE=~5.34ms, TI=~1100ms, 1mm isotropic voxel size).

\begin{figure}[!t]
	\centering
	\includegraphics[width=0.4\textwidth]{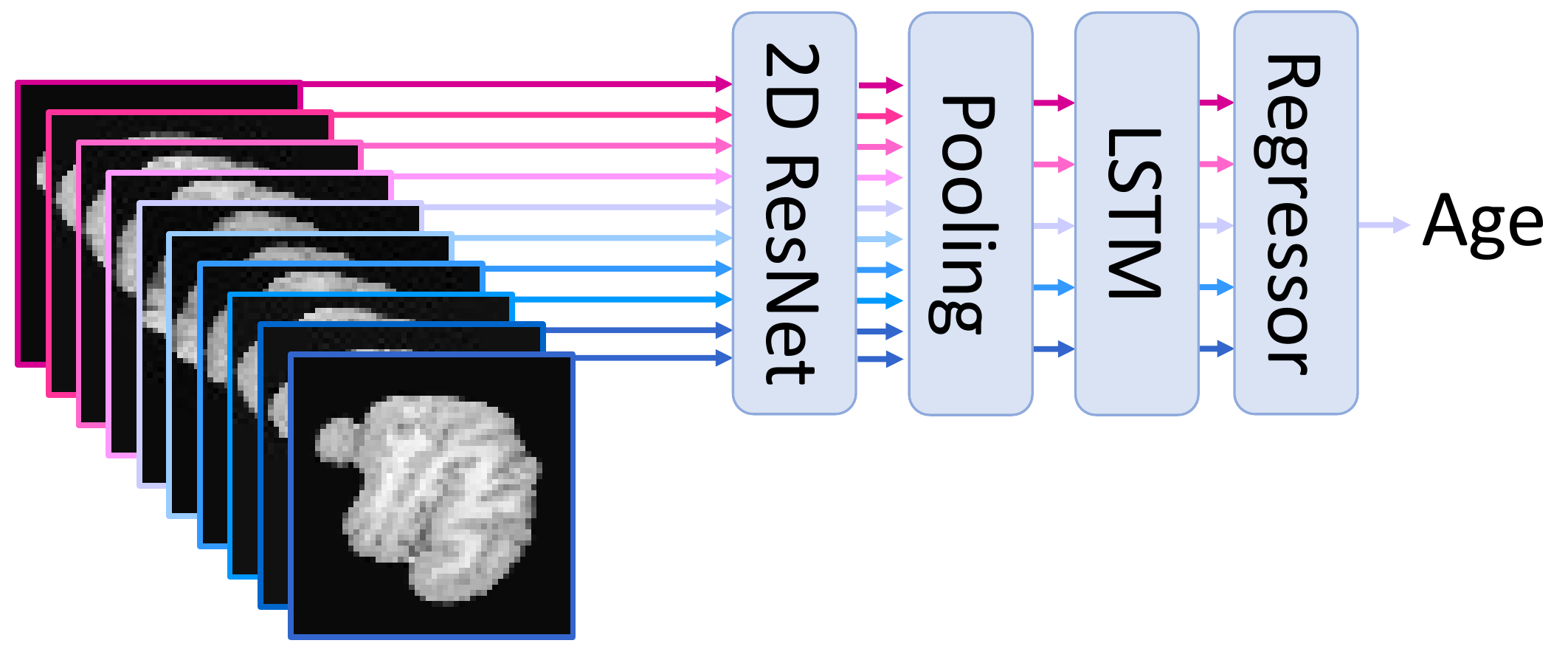}
	\caption{Illustration of the 2D-ResNet with LSTM framework for age prediction. The 3D MRI is considered as a sequence of 2D images. The 2D-Resnet is used to extract features and then followed by a pooling operation. LSTM is used to capture the contextual information over the sequence and the final regressor is applied to estimate the age. }
	\label{fig:reslstm}
\end{figure}

\section{Methods}
\label{sec:method}

In this section, we describe the details of the proposed method, named as 2D-ResNet18+LSTM, which consists of four main parts (shown in Fig.~\ref{fig:reslstm}): feature extraction by 2D residual network with 18 layers,
a pooling for feature reduction, LSTM for context modeling over the sequence and a final regressor for age predication.

\textbf{Feature extraction.}
Given a sequence of 2D slices of a 3D brain MRI volume, $\textbf{X}=\{x_1,x_2,\cdots,x_n\}$ (where $n$ is the number of slices), 2D-ResNet~\cite{he2016deep} is used to extract feature representation.
In this paper, we use ResNet18 without the fully-connected layer as the backbone to extract feature $f_t$ for each 2D image $x_t$:
\begin{equation}
f_t = \text{ResNet}(\Theta,x_t)
\end{equation}
where $f_t$ is the feature representation of $x_t$ and $\Theta$ is the set of parameters for ResNet18.
We use ReLU after each convolutional layer as the activation function.
%Finally, the dimension of $f_t$ is 512.

\textbf{Pooling.}
Looking at 2D slices one by one at a constant pace may not be an optimal way to gather age information. It is likely that pooling information from several adjacent 2D slices can smooth out noises in single slices and can better reveal true age information. This motivates us to combine features across adjacent 2D slices by a pooling operation. The most common pooling method is the average pooling, which is defined as:
\begin{equation}
p(m)_{avg} = \frac{1}{k}\sum_{j=k*m,...,k*m+k} f_j
\end{equation}
where $k$ is the kernel size of the pooling.
Finally, the sequence is reduced to $P=\{p_1,p_2,\cdots,p_m\}$ where $m=n/k$.
In the experiments, we set $k=3$.

\textbf{LSTM.}
As mentioned above, each 3D brain MRI can be considered as a sequence of 2D slices, which, after pooling, are in $m$ stacks of 2D slices represented by the feature vector $P=\{p_1,p_2,\cdots,p_m\}$. This formulation motivates us to use recurrent neural networks, such as LSTM~\cite{hochreiter1997long}, to capture the age information.
The LSTM cell (shown in Fig.~\ref{fig:lstmcell}) is defined as:

\begin{figure}[!t]
	\centering
	\includegraphics[width=0.25\textwidth]{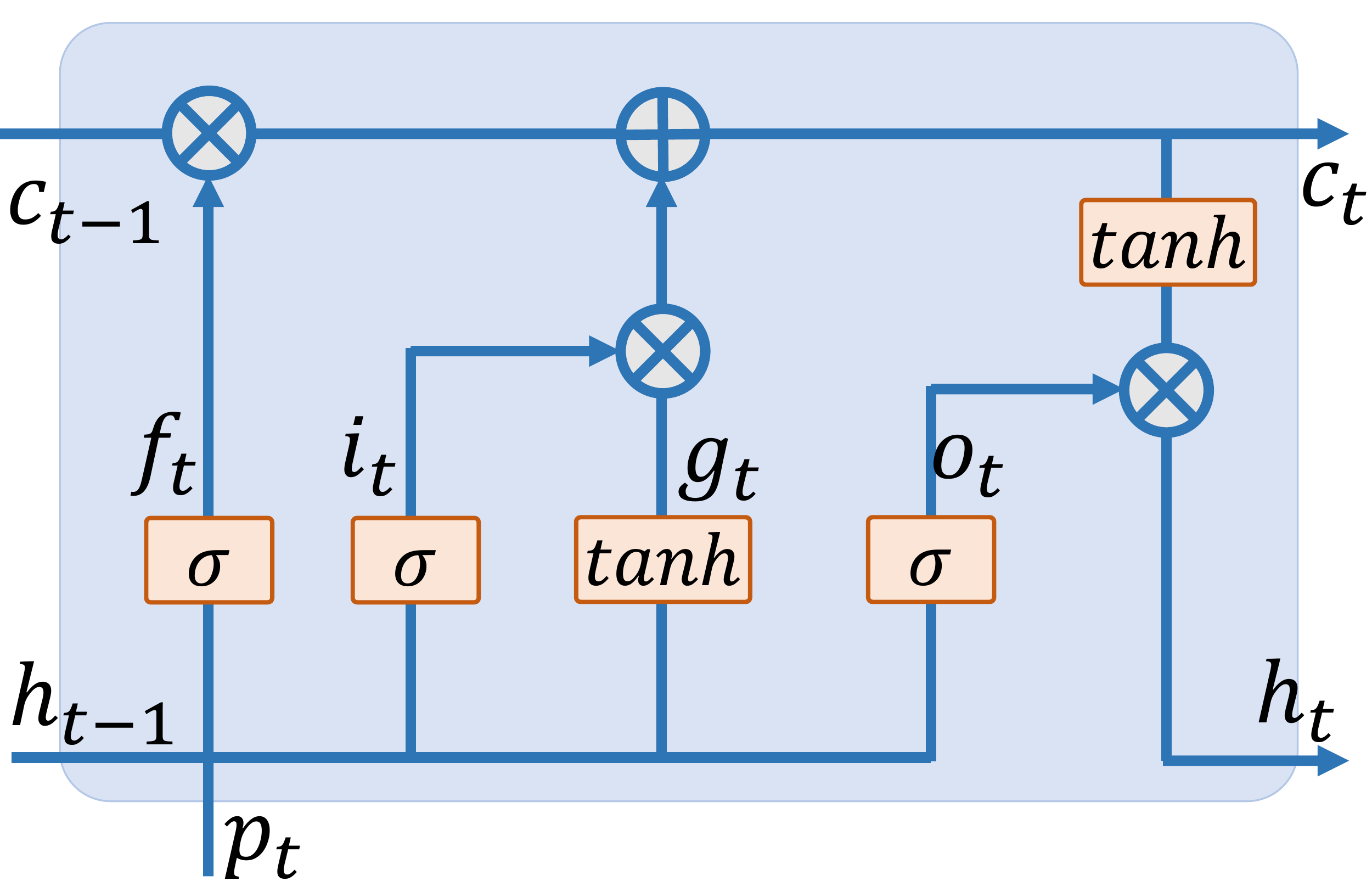}
	\caption{The structure of LSTM cell, which consists of a cell module $c_t$, a hidden state $h_t$, a forget gate $f_t$, a input gate $i_t$ and a output gate $o_t$. }
	\label{fig:lstmcell}
\end{figure}

\vspace{-0.5cm}
\begin{equation}
\begin{split}
i_t &= \sigma(W_{ix}\cdot p_t + W_{ih} \cdot h_{t-1} + b_i) \\
f_t &= \sigma(W_{fx}\cdot p_t + W_{fh} \cdot h_{t-1} + b_f) \\
o_t &= \sigma(W_{ox}\cdot p_t + W_{oh} \cdot h_{t-1} + b_o) \\
g_t &= \varphi(W_{gx}\cdot p_t + W_{gh} \cdot h_{t-1} + b_g) \\
c_t &= f_t \odot c_{t-1} + i_t \odot g_t \\
h_t &= o_t \odot \varphi (c_t) \\
\end{split}
\end{equation}
where $\sigma(\cdot)$, $\varphi(\cdot)$ are the sigmoid and tangent $tanh(\cdot)$ function, respectively. $\odot$ denotes an element-wise product.
$W_{*x}$, $W_{*h}$ and $b_*$ are parameters which are learned during training. 
We use the Bidirectional LSTM with the hidden dimension size of 64, to balance between the performance and the computing cost.

\textbf{Age regressor.} All the outputs of LSTM cells are concatenated as the feature vector $\vec{H}=[h_1,h_2,...,h_t]$ and a fully connected layer is used as the regressor for age prediction.

\textbf{Instance Normalization (IN).} Use 2D/3D batch normalization when the batch size is small (such as 1) may quickly explode the computer memory. Therefore, in this paper, we resort to instance normalization~\cite{ulyanov2016instance}, which is defined as:
\begin{equation}
    y = \frac{x-\text{E}[x]}{\sqrt{\text{Var}[x]+\epsilon}}
\end{equation}
where the input $x$ is a 2D/3D feature map. $\text{E}$ and $\text{Var}$ are the mean and variation, respectively.
Instance Normalization is applied after each 2D/3D convolutional layer.

\section{Experimental results}

We use the NIH-PD dataset as the Discovery Cohort. We partitioned the 1212 subjects (0-22 years of age at the time of MRI scans) in the NIH-PD data into 
training (80\%) and testing sets (20\%) to build the model and evaluate the model performance during the Discovery phase. Then, we use the MGHBCH dataset, which contains 428 normal-developing subjects 0-6 years of age, as a totally "unseen" Replication Cohort.

Our second setting to test the generalization ability of age prediction in the MGHBCH Replication Cohort (0-6 years of age) is to compare the accuracies by training on NIH-PD subjects 0-22 years and by training on NIH-PD subjects 0-6 years of age. The age distribution of each dataset is shown in Fig.~\ref{fig:model}.

T1-weighted MPRAGE MRI of every subject is registered to the SRI atlas (image size [128,160,112] and voxel size $1\times1\times1 mm$ \cite{rohlfing2010sri24}). We compared using the SRI atlas (adults) and a 2-year-old pediatric atlas and found little differences in the final accuracies. We resize 2D images on the sequence to 50$\times$50 voxels to reduce the computation time and the use of memory.
The voxel values of each subject are normalized by subtracting the mean value of the subject's image and dividing by the variance.

%The backbone of the proposed 2D-ResNet18-LSTM is the ResNet18~\cite{he2016deep} with 2D images as input to extract deep features.
We use the corresponding 3D-ResNet18 as the baseline for comparison, which consists of 3D-convolutional layers with kernel size of 3$\times$3$\times$3, global average pooling and a fully connected layer for age prediction.
All networks, including 2D-ResNet18+LSTM and 3D-ResNet18 are trained with Mean Absolute Error (MAE) loss. 
The Adam optimization method is used with an initial learning rate of 0.0001, decreasing by half at every 15 epochs.
The network is trained with 60 epochs. The batch size is set to 1.
%As mentioned above, the instance normalization is applied after each 2D/3D convolutional layer.

The model is evaluated using two measurement metrics: Mean Absolute Error (MAE) and Cumulative Score (CS):
\begin{equation}
    \begin{split}
        MAE &= \frac{1}{N} \sum_{i=1}^N |y_i-\hat{y_i}| \\
        CS(\alpha) &= \frac{1}{N} \sum_{i=1}^N \Big\{ |y_i-\hat{y_i}| \leq \alpha \Big\} \times 100\%
    \end{split}
\end{equation}
where $y_i$ is the ground-truth age, $\hat{y_i}$ is the estimated age, $\alpha$ is the error level and $N$ is the number of test samples.
The CS$(\alpha)$ score is the percentage of subjects whose errors of age prediction are smaller than or equal to $\alpha$. A smaller $MAE$ and a higher $CS(\alpha)$ indicate a more accuracy algorithm.

\subsection{Performance on the NIH-PD testing set (Discovery)}

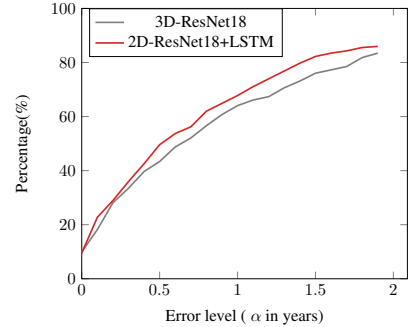
\begin{figure}[!t]
    \centering
    \resizebox{0.3\textwidth}{!}{
			    \begin{tikzpicture}
			    
                    \begin{axis}[
                    	%x tick label style={/pgf/number format/0.1 sep=},
                    	%xtick={0,0.1,0.2,0.3,0.4,0.5,0.6},
                    	ylabel=Percentage(\%),
                    	xlabel=Error level ( $\alpha$ in years),
                    	ymax=100,ymin=0,
                    	legend style={at={(0.32,1.0)},
                    	anchor=north,legend columns=1},
                    	%ybar interval=0.3,
                    	%width=8cm,
                    	xmin=0,
                    ]
                    \addplot[line width=1pt,gray,mark='']
                    	coordinates {(0.00,9.92)(0.10,18.18)(0.20,28.10)(0.30,33.47)(0.40,39.67)(0.50,43.39)(0.60,48.76)(0.70,52.07)(0.80,56.61)(0.90,60.74)(1.00,64.05)(1.10,66.12)(1.20,67.36)(1.30,70.66)(1.40,73.14)(1.50,76.03)(1.60,77.27)(1.70,78.51)(1.80,81.82)(1.90,83.47)};
                  
                    \addplot[line width=1pt,red!70!gray,mark='']
                    	coordinates {(0.00,9.50)(0.10,22.73)(0.20,28.93)(0.30,35.95)(0.40,42.56)(0.50,49.59)(0.60,53.72)(0.70,56.20)(0.80,61.98)(0.90,64.88)(1.00,67.77)(1.10,71.07)(1.20,73.97)(1.30,76.86)(1.40,79.75)(1.50,82.23)(1.60,83.47)(1.70,84.30)(1.80,85.54)(1.90,85.95)};
                    	
                    %\addplot coordinates {(1,0.39)(3,0.3639) (5,0.439)(7,0.45)};
                    \legend{3D-ResNet18,2D-ResNet18+LSTM}
                    \end{axis}
                    %\draw[gray,dashed,line width=2pt](3.3,0)--(3.3,4.5);
                \end{tikzpicture}}
    \caption{Cumulative score (CS) on the error levels from 0 to 2 years of different methods on the Discovery Cohort (NIH-PD testing set). A higher CS is better.}
    \label{fig:csnihpd}
\end{figure}

Table~\ref{tab:nihpdmae} shows the MAEs of different methods in different age groups on the NIH-PD testing set.
The proposed 2D-ResNet18+LSTM provides lower MAEs in 6-10, 11-15 and 16-22 years, and gives a lower average MAEs than 3D-ResNet18.
Fig.~\ref{fig:csnihpd} shows the CS scores as a function of error levels $\alpha$.
%and Table~\ref{tab:csnihpd} shows the MAEs and CS with error level $\alpha=1.0$ year of different methods.
The proposed 2D-ResNet18+LSTM provides better results than 3D-ResNet18 in terms of MAE and CS.

\begin{table}[!t]
    \centering
    \caption{Accuracy in the Discovery Cohort. The MAE (in years) for brain age estimation on the NIH-PD testing set. Numbers in bold are more accurate results.}
    \label{tab:nihpdmae}
    \resizebox{0.48\textwidth}{!}{
    \begin{tabular}{l|cccc|c}
    \hline 
    \multirow{2}{*}{Method} & \multicolumn{4}{c|}{Age group } & \multirow{2}{*}{Average} \\
    \cline{2-5}
         & 0-5 & 6-10 & 11-15 & 16-22 & \\
        \hline 
        3D-ResNet18 & \textbf{0.30} & 1.12 & 1.19 & 1.85 & 1.07 \\
        2D-ResNet18+LSTM & 0.32 & \textbf{0.92} & \textbf{1.02} & \textbf{1.79} & \textbf{0.96}\\
        \hline 
    \end{tabular}}
\end{table}

%\begin{table}[!t]
%	\centering 
%	\caption{Cumulative Score (SC) of different methods with error level %$\alpha=1.0$ years on the NIH-PD test dataset.}
%	\label{tab:csnihpd}
%	\begin{tabular}{l|cc}
%    \hline 
%        Method & MAE &  CS($\alpha$=1.0 years) \\
%       \hline 
%       3D-ResNet18 & 1.07 & 64.05\% \\
%       2D-ResNet18+LSTM & \textbf{0.96} & \textbf{67.77\%}\\
%       \hline 
%   \end{tabular}
%\end{table}

\subsection{Generalization/Replication to the MGHBCH dataset}

In this section, we provide the results of age estimation on the MGHBCH dataset using the model trained on the NIH-PD training set. This is to evaluate the generalization of 2D-ResNet18+LSTM and 3D-ResNet18 methods.

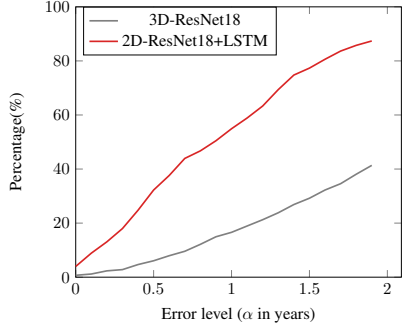
\begin{figure}[!t]
    \centering
    \resizebox{0.3\textwidth}{!}{
			    \begin{tikzpicture}
			    
                    \begin{axis}[
                    	%x tick label style={/pgf/number format/0.1 sep=},
                    	%xtick={0,0.1,0.2,0.3,0.4,0.5,0.6},
                    	ylabel=Percentage(\%),
                    	xlabel=Error level ($\alpha$ in years),
                    	ymax=100,ymin=0,
                    	legend style={at={(0.32,1.0)},
                    	anchor=north,legend columns=1},
                    	%ybar interval=0.3,
                    	%width=8cm,
                    	xmin=0,
                    ]
                    \addplot[line width=1pt,gray,mark='']
                    	coordinates {(0.00,0.70)(0.10,1.17)(0.20,2.34)(0.30,2.80)(0.40,4.67)(0.50,6.07)(0.60,7.94)(0.70,9.58)(0.80,12.15)(0.90,14.95)(1.00,16.59)(1.10,18.93)(1.20,21.26)(1.30,23.83)(1.40,26.87)(1.50,29.21)(1.60,32.24)(1.70,34.58)(1.80,38.08)(1.90,41.36)};
                  
                    \addplot[line width=1pt,red!70!gray,mark='']
                    	coordinates {(0.00,3.97)(0.10,8.88)(0.20,13.08)(0.30,17.99)(0.40,24.77)(0.50,32.24)(0.60,37.62)(0.70,43.93)(0.80,46.73)(0.90,50.47)(1.00,54.91)(1.10,58.88)(1.20,63.32)(1.30,69.39)(1.40,74.77)(1.50,77.34)(1.60,80.61)(1.70,83.64)(1.80,85.75)(1.90,87.38)};
                    	
                    %\addplot coordinates {(1,0.39)(3,0.3639) (5,0.439)(7,0.45)};
                    \legend{3D-ResNet18,2D-ResNet18+LSTM}
                    \end{axis}
                    %\draw[gray,dashed,line width=2pt](3.3,0)--(3.3,4.5);
                \end{tikzpicture}}
    \caption{Cumulative score (CS) on the error levels from 0 to 2 years trained on the Discovery Cohort (NIH-PD training set) and applied to the Replication Cohort. A high CS is better.}
    \label{fig:csmghbch}
\end{figure}

Table~\ref{tab:mgmea} shows the MAEs of these two methods on the MGHBCH dataset (0-6 years) after training on NIH-PD subjects 0-22 year of age.
The proposed 2D-ResNet18+LSTM provides much better results than 3D-ResNet18 on the MGHBCH dataset. Fig.~\ref{fig:csmghbch} shows the CS curves of these two methods and the proposed 2D-ResNet18+LSTM provides better results than 3D-ResNet18 with all different error levels $\alpha$.

\begin{table}[!t]
    \centering
    \caption{Accuracy in the Replication Cohort, training on Discovery Cohort 0-22 years of age. The MAE (in years) of different methods for brain age estimation trained on the NIH-PD training set and replicated on the truly "unseen" MGHBCH dataset for examining the generalization ability. Numbers in bold are more accurate results.}
    \label{tab:mgmea}
    \resizebox{0.48\textwidth}{!}{
    \begin{tabular}{l|cccccc|c}
    \hline 
    \multirow{2}{*}{Method} & \multicolumn{6}{c|}{Age group} & \multirow{2}{*}{Average} \\
    \cline{2-7}
         & 0-1 & 1-2 & 2-3 & 3-4 & 4-5 & 5-6  \\
        \hline 
        3D-ResNet18 &  2.57 & 2.43 & 2.80 & 2.64 & 2.85 & 2.75 & 2.64\\
        2D-ResNet18+LSTM & \textbf{0.83} & \textbf{1.65} & \textbf{1.40} & \textbf{1.36} & \textbf{1.21} & \textbf{0.83} & \textbf{1.14}\\
        \hline 
    \end{tabular}}
\end{table}

%Fig.~\ref{fig:scatter} shows the scatter plot of the estimated age and chronological age. 
%Most estimated ages of 3D-ResNet18 are larger than their chronological age
%while the estimated ages of the proposed method are close to their chronological age, yielding a lower MAE.

%Table~\ref{tab:csmghbch} provides the MAEs and CSs with error level $\alpha=1.0$ year.
%The estimated errors by the proposed 2D-ResNet18+LSTM of around 54.91\% subjects on the MGHBCH dataset are less than or equal to 1 year which is higher than 16.59\% given by the 3D-ResNet18 network.
%Around 54.91\% subjects whose prediction errors given by the 2D-ResNet18+LSTM are less than 1 year, which is higher than 16.59\% sujects given by the the 3D-ResNet18 network.

%\begin{table}[!t]
%	\centering 
%	\caption{Cumulative Score (SC) of different methods with error level %$\alpha=1.0$ years  trained on the NIH-PD training set and tested on the %unseen MGHBCH dataset for generalization ability.}
%	\label{tab:csmghbch}
%	\begin{tabular}{l|cc}
%    \hline 
%        Method & MAE &  CS($\alpha$=1.0 years) \\
%        \hline 
%        3D-ResNet18 & 2.64 & 16.59\% \\
%        2D-ResNet18+LSTM & \textbf{1.14} & \textbf{54.91\%}\\
%        \hline 
%   \end{tabular}
%\end{table}

%From the experimental resulrs we can see that the proposed 2D-CNN+LSTM can provide better generalization results than 3D-CNN 
%because 3D-ResNet18 integrates the 3-dimensional information on every stages by 3D convolutional operations which is sensitive to input noise or bias introduced by the difference between different site of datasets.
%However, 2D-ResNet18+LSTM integrates the temporal information on the high-level feature space of 2D images which is less sensitive to the input noise or bias.

\begin{table}[!t]
    \centering
    \caption{Accuracy in the Replication Cohort, training on Discovery Cohort 0-6 years of age. The MAE (in years) of different methods for brain age estimation trained on the NIH-PD training set with subjects of age 0-6 years and replicated on the truly "unseen" MGHBCH dataset for examining the generalization. Numbers in bold are more accurate results.}
    \label{tab:smallset}
    \resizebox{0.48\textwidth}{!}{
    \begin{tabular}{l|cccccc|c}
    \hline 
    \multirow{2}{*}{Method} & \multicolumn{6}{c|}{Age group} & \multirow{2}{*}{Average} \\
    \cline{2-7}
         & 0-1 & 1-2 & 2-3 & 3-4 & 4-5 & 5-6  \\
        \hline 
        3D-ResNet18 &  0.66 & \textbf{0.48} & \textbf{0.43} & \textbf{0.75} & 1.26 & 2.10 & 0.79 \\
        2D-ResNet18+LSTM & \textbf{0.24} & 1.34 & 1.43 & 1.25 & \textbf{0.69} & \textbf{0.92} & \textbf{0.78}\\
        \hline 
    \end{tabular}}
\end{table}

We repeat the above experiment in the Discovery (MGHBCH) Cohort, but using NIH-PD subjects 0-6 years for training. Table~\ref{tab:smallset} shows the results.
When the Discovery and Replicate Cohorts share the same age range, the average MAEs of 2D-ResNet18+LSTM and 3D-ResNet18 are approximately the same, but the computation time of 2D-ResNet18+LSTM is shorter than 3D-ResNet18 (3 hours versus 7 hours on an NVIDIA K80 GPU).

\section{Conclusion}
Recent age prediction is mainly on 18-90 years of age and using 2000-3000 brain MRIs. We focus on 0-22 especially 0-6 years where the sample size is smaller. In such circumstances, we shown that the proposed 2D-ResNet18+LSTM method provides comparative or better results than 3D-ResNet18.

While existing age prediction studies split the Discovery Cohort into training and testing, we further test the generalization ability in a truly unseen Replication Cohort (subjects unseen, imaging sites/protocols unseen). We show that the proposed 2D-ResNet18+LSTM method generalizes better to the Replication dataset when the age ranges of Discovery and Replication Cohorts are different (Table~\ref{tab:mgmea}). This opens a window to train on a large-scale Discovery Cohort with a wider age range (such as 0-20 years) and to adapt to a Replication Cohort where the age range of interest is narrower (0-2 or 0-6 years) and the sample size is smaller.

When the age ranges of Discovery and Replication Cohorts are the same, 2D-ResNet18+LSTM has a faster computation, and offers a higher accuracy in both ends of the age range (Table~\ref{tab:smallset}) while 3D-ResNet18 has a higher accuracy in the middle of the age range. This suggests us to explore ways to combine the proposed 2D-ResNet18+LSTM and the 3D-ResNet18 methods to achieve high accuracies throughout the age range of interest. Our future work also includes more comprehensive evaluation in large multisite datasets and application to probing brain disorders in early life.

% References should be produced using the bibtex program from suitable
% BiBTeX files (here: strings, refs, manuals). The IEEEbib.bst bibliography
% style file from IEEE produces unsorted bibliography list.
% -------------------------------------------------------------------------
\bibliographystyle{IEEEbib}

\bibliography{IEEEabrv,refs}

\end{document}